\documentclass[letterpaper]{article} 
\usepackage{aaai23}  
\usepackage{times}  
\usepackage{helvet}  
\usepackage{courier}  
\usepackage[hyphens]{url}  
\usepackage{graphicx} 
\usepackage{natbib}  
\usepackage{caption} 
\usepackage{algorithm}
\usepackage{algorithmic}

\usepackage{newfloat}
\usepackage{listings}
\DeclareCaptionStyle{ruled}{labelfont=normalfont,labelsep=colon,strut=off} 
\floatstyle{ruled}
\newfloat{listing}{tb}{lst}{}
\floatname{listing}{Listing}
\title{Planning for Learning Object Properties}
\author {
    Leonardo Lamanna,\textsuperscript{\rm 1,\rm 2}
    Luciano Serafini, \textsuperscript{\rm 2}
    Mohamadreza Faridghasemnia, \textsuperscript{\rm 3}
    Alessandro Saffiotti, \textsuperscript{\rm 3}
    Alessandro Saetti \textsuperscript{\rm 2}
    Alfonso Gerevini \textsuperscript{\rm 2}
    Paolo Traverso \textsuperscript{\rm 1}
}
\affiliations {
    \textsuperscript{\rm 1} Fondazione Bruno Kessler, Trento, Italy\\
    \textsuperscript{\rm 2} Department of Information Engineering, University of Brescia, Italy\\
    \textsuperscript{\rm 3} Center for Applied Autonomous Sensor Systems, University of \"{O}rebro, Sweden\\
    llamanna@fbk.eu, serafini@fbk.eu, mohamadreza.farid@oru.se,  asaffio@aass.oru.se, alessandro.saetti@unibs.it, alfonso.gerevini@unibs.it, traverso@fbk.eu
}

\usepackage{bm}
\usepackage{colortbl}
\usepackage{tikz}
\usetikzlibrary{shapes}
\usepackage{caption}
\usepackage{subcaption}
\usepackage{multirow}
\usepackage{url}
\usepackage{refcount}
\usepackage{amsfonts}
\usepackage{comment}
\usepackage{booktabs}
\usepackage{amsmath}
\usepackage{verbatimbox}

\def\planprob{\Pi}

\def\C{\mathcal{C}}
\def\H{\mathcal{H}}

\def\P{\mathcal{P}}
\def\OP{\mathcal{O}}

\def\O{\mathcal{O}}

\def\D{\mathcal{D}}

\def\eff{\mathsf{eff}}
\def\prec{\mathsf{pre}}
\def\param{\mathsf{par}}

\def\code#1{{\texttt{\footnotesize{#1}}}}
\def\name#1{\textrm{`}#1\textrm{'}}

\begin{document}

\maketitle

  \begin{abstract}
Autonomous agents embedded in a physical environment need the ability to recognize objects and their properties from sensory data.
Such a perceptual ability is often implemented by supervised machine learning models, which are pre-trained using a set of labelled data.
In real-world, open-ended deployments, however, it is unrealistic to assume to have a pre-trained model for all possible environments. Therefore, agents need to dynamically learn/adapt/extend their perceptual abilities online,
in an autonomous way, 
by exploring and interacting with the environment where they operate.
%
%
This paper describes a way to do so, by exploiting symbolic planning. Specifically,
we formalize the problem of automatically training a neural network to recognize
object properties 
as a symbolic planning problem 
(using PDDL). 
We use planning techniques to produce a strategy for automating the training dataset creation and the learning process.
%
Finally, we provide an experimental evaluation in both a simulated and a real environment, which 
shows that the proposed approach is able to successfully learn how to recognize new object properties.
\end{abstract}

\section{Introduction}

Agents embedded in a physical environment, like autonomous robots,
need the ability to perceive objects in the environment and recognize
their properties. For instance, a robot operating in an indoor environment 
should be able to recognize whether a certain box found in the environment 
is open or closed. 
From these perceptions the agent can build and use abstract 
representations of the states of the environment to reach its goals
by automatic planning techniques.
%
The common approach to provide an agent with such perceptual
capabilities consists in pre-training offline a (set of) perception models
from hundreds of thousands of 
annotated data 
(e.g., images or other sensory data). See for instance 
\cite{asai2019,dengler2021online, lamanna2022online}.


In offline training approaches, the perception capabilities are fixed
once and for all. This is in stark contrast with a main requirement in
many robotics applications: agents embedded in real-world, open-ended
environments should be able to dynamically and autonomously improve
their perceptual abilities by actively exploring their environments.
This is also in agreement with the emerging popular research area of
interactive perception \cite{bohg2017}.  When perception functions are
modelled by (deep) neural networks, an open and interesting challenge is
whether agents can autonomously decide when and how to improve their perception
models, by collecting the needed training data and using them to train the
neural network.


In this paper, we explore a way to address this challenge with an
automated planning approach. In particular, we design a PDDL planning domain
\cite{McDermott:98-PDDL} for 
\emph{planning to
  learn (or improve) the perceptual capabilities of the agent}. We focus on the
problem of automatically training neural networks able to recognize
properties of objects, e.g., open/closed, by relying on a pre-trained object detector.
We extend a PDDL planning domain, called \emph{base domain}, 
with new actions and predicates for learning properties of object types. 
Such an extension, called \emph{learning domain}, is specified in a meta-language of
the base language. It contains the reification of 
properties and types of the base language. For instance, if 
\code{Is\_Open} is a property of the base domain, the learning domain 
contains the object \name{is\_open} of type \code{Property}. 
Furthermore, the learning domain contains actions for collecting 
training examples for properties,  and actions for training the 
network with them.

The online learning of object properties is obtained by planning in 
the union of the base and learning domains, and by executing the 
generated plan. The base domain allows the agent to 
plan to reach a state where the agent can observe an object with a certain property,
e.g., a state where \code{Is\_Open(box$_0$)} is true. The learning domain allows the 
agent to plan for actions that collect observations of objects with the property being true, 
e.g., take pictures of \code{box$_0$},  which is known to be open, and add them
to the positive training examples for the property \code{Is\_Open}.
In this way, the agent automatically maps low level perceptions (e.g., 
images of an open box) into the symbolic property of objects at the 
abstract planning level (e.g., ``the box is open" in PDDL). 


We provide an experimental evaluation both in a photo-realistic simulated
environment \cite{ai2thor} and in a real world setting. In the
simulated environment, we evaluate the ability to learn certain object
properties
under two different
conditions of object detection: noisy and ground truth. The results
indicate that the approach is able to successfully learn and recognize
the object properties with high precision/recall for most object
classes. In the real world environment, actions for changing object
properties (e.g., opening a book) are performed cooperatively
with humans, e.g., by asking a human to
physically perform the required action.



\section{Related Work}
\label{sec:relatedwork}

As far as we know, the proposed approach is new.
Several approaches are based on the idea of exploiting automated learning for different planning tasks, like for learning action models, heuristics, plans, or policies.
Most of the research on learning action models, 
see, e.g.,  \cite{aineto2018learning,aineto2019learning,bonet2020,mccluskey2009,cresswell2013,gregory2016,lamanna2021IJCAI,juba2021}, does not deal with the problem of learning from real value perceptions.
%
Other works have addressed the problem of learning planning domains
from perceptions in the form of high dimensional raw data (such as images), 
see, e.g., \cite{asai2018,asai2019,janner2018reasoning,
dengler2021online,
konidaris2018,
liberman2022learning}.
%
In these works, the abstract planning domain is 
obtained by offline pre-training, and the mapping between perceptions to the abstract model is fixed, while we learn/adapt this mapping online.
%
%
Our approach shares some similarities with the work on planning by reinforcement learning (RL)
\cite{sutton98},
since we learn by acting in a (simulated) environment, and especially with the work on deep RL (DRL) \cite{mnih2015,mnih2016},
which dynamically trains a deep neural network by interacting with the environment. 
%
However, DRL focuses on learning policies and perceptions are mapped into state embeddings that cannot be easily mapped into human comprehensible symbolic (PDDL) states.
In general, while all the aforementioned works address the problem of using learning techniques for planning,
we address a different problem, i.e., using planning techniques for learning 
automatically and online
to recognize object properties from low level high dimensional data.

We share a similar motivation of the research on interactive perception, see \cite{bohg2017} for a comprehensive survey, and especially the work in this area for learning properties of objects, see, e.g., \cite{natale2021}. However, most of this work has the objective to integrate acting and learning, and to study the relation between action and sensory response.
We instead address the challenge of building autonomous systems with planning capabilities that can automate the training and learning process.

A wide variety of approaches and applications have been proposed for enhancing robotic agents with active learning techniques \cite{kulick2013active,cakmak2012designing,cakmak2010designing,chao2010transparent,ribes2015active,hayes2014discovering,huang2010active,ashari2019mindful}. In these works, the robotic agents improve their skills or learn new concepts by collecting and labeling data in an online way. However, all these methods label data by means of either human supervisions or a confidence criteria applied on the prediction of a pre-trained model. In contrast, our approach does not require human supervision, and collected data are labelled by applying actions.

In \cite{ugur2015bottom}, similarly to our approach, the authors propose a method for learning the predicates corresponding to action effects after their executions. They learn action effects by clustering hand-crafted visual features of the manipulated objects, extracted from the continuous observations obtained after executing the actions. However, the learned predicates lack of interpretation, which must be given by a human. On the contrary, our approach learns explainable predicates.
Moreover, they focus on learning the effects of a single action (i.e., stacking two blocks) in a fully observable environment using ground truth object detection. Our approach learns predicates corresponding to the effects of several actions in partially observable environments, and using a noisy object detector.

The approach by \cite{migimatsu2022grounding} learns to map images into the truth values of predicates of planning states. Differently from us, their approach is offline and requires the sequence of images labeled with actions, while our approach plans for generating this sequence online.
We share the idea of learning state representations through interaction with \cite{pinto2016curious}, where they learn visual representations of an environment by manipulating objects on a table. Notably, they learn the visual representation in an unsupervised way, through a CNN  trained on a dataset generated by interacting with objects. However, the learned representations lack of interpretation. Furthermore, in \cite{pinto2016curious}, the learned representations are not suitable for applying symbolic planning. 

Finally, our extension of the planning domain with a meta-language shares some commonalities with the work on planning at the knowledge level 
(see, e.g., \cite{petrick02}), which addresses the different problem of planning with incomplete information and sensing.

\section{Preliminaries}
\label{sec:preliminaries}


\paragraph{Symbolic planning}
A planning domain $\D$ is a tuple $\left<\P, \OP,\H\right>$ where $\P$ is a set of
first order predicates with associated arity, $\OP$  
is a set of operators with associated arity, and
$\H$ associates to every operator $op\in\OP$
an action schema. The action schema is composed of a triple $\left<\param(op),\prec(op),\eff^+(op),\eff^-(op)\right>$ in which
$\param(op)$ is a $n$-tuple of distinct variables $x_1,\dots,x_n$,
where $n$ is the arity of $op$, and $\prec(op)$
is a set of first order formulas with predicates in $\P$ and
arguments in $x_1,\dots,x_n$, and 
$\eff^+(op)$ and $\eff^-(op)$
are set of \emph{atomic} first order formula with predicates in $\P$ and
arguments in $x_1,\dots,x_n$. 
Among the unary predicates of $\P$, we distinguish between object types
and object properties, which are denoted with $t$ and $p$, respectively. 

Given a planning domain $\D$ and a set of constants $\C$, a grounded planning
domain is obtained by grounding all the actions schema of $\D$ with the constants in $\C$. 
The set of states of a grounded planning domain $\D(\C)$
is the set of all possible subsets of atoms that can be built 
by instantiating every predicate in $\P$ in all possible ways with the constants in $\C$.
A ground action model defines a transition function
among states where $(s,op(c_1,\dots,c_n),s')$ if and only if $s\models\prec(op(c_1,\dots,c_n))$
and 
$s' = s\cup\eff^+(op(c_1,\dots,c_n))\setminus\eff^-(op(c_1,\dots,c_n))$.

A planning problem $\planprob$ on a grounded planning domain $\D(\C)$
is a triple $\planprob=\left<\D(\C),s_0,g\right>$, where $s_0$ is an initial state
and $g$ is a 
first order formula that identifies a set of states of $
D(\C)$ in which the goal is satisfied, i.e.,
$\{s\in 
2^{\P(\C)}\mid s\models g\}$.
A plan $\pi$ for problem $\planprob$ is a sequence of ground actions
$\langle a_1,\dots,a_k \rangle$ such that $(s_{i-1},a_{i},s_{i})$ for $i=1,\dots,k$ is a transition
of $\D(\C)$ and $s_{k}\models g$. 

\paragraph{Perception functions}
The agent perceives the environment by sensors that return real-value 
measurements
on some portion of the environment.
For example, the perception of an agent with a on-board camera and a system for estimating its position
consists in a vector $(x,y,z)$ of coordinates and an RGB-D image taken by the agent's on-board camera.
Observations are partial (e.g., the camera provides only the front view) and could be incorrect (e.g.,
the estimation of the position could be noisy). 
We suppose that, at the time when 
a perception occurs, 
the agent's knowledge about the environment is represented by a grounded planning domain $\D(\C)$, where $C$ represents the set of objects already discovered in the environment.
Each $c\in\C$ is associated with an anchor \cite{coradeschi2003introduction} that describes the perceptual features of $c$ that have been collected by the agent so far 
(e.g., the pictures of $c$ from different angles, the estimated position and size of $c$, etc.).
At the beginning, the set $\C$ of constants is empty. 
The agent performs and processes each perception in order to extract some knowledge about the objects 
in the environment, and about their properties in the current state. 
This is achieved by combining an \emph{object detector} and a set of
\emph{property classifiers}.

The object detector identifies a set of
objects in the current perception (e.g., RGB-D image) and predicts
their types (i.e., it selects one type among the object types of the
planning domain). Every detected object is associated with numeric
features (e.g., the bounding box, the estimation of the position, etc.),
which are used to build the anchors of the detected object. The
features of each detected object are compared with the features of the
objects already known by the agent, i.e., those present in the current
set of constants $\C$. 
If the features of the detected object matches
(to a certain degree) the features of a $c\in\C$, then the features of
$c$ are updated with the new discovered features. Otherwise $\C$ is
extended with a new constant $c$ anchored to the features of the
detected object, and the type $t(c)$ is asserted in the planning
domain, where $t$ is the type returned by the object detector.

For every object $c$ of type $t$ returned by the object detector and 
for every property $p$ that applies to $t$,
a classifier $\rho_{t,p}$ predicts if $c$ has/has not the property $p$.
Notice that not all properties apply to a type,
e.g., a laptop cannot be filled or empty.
Furthermore, for the same property we use different classifiers
for different types, since predicting that a bottle is open or that a book is open
from visual features are two very different tasks. 
$\rho_{t,p}$ can be specified either explicitly by a set of predefined rules,
or it can be a machine learning model trainable by supervised examples. 
For instance, the classifier that checks if an object is \code{Close\_To} the agent is defined
by a threshold on the distance between the agent and the object position. 
Other properties (e.g., \code{Is\_Open}) are predicted using a neural network,
which takes as input object images and returns the probability
of the property being true. 

%
%

%


\paragraph{Plan execution}
To achieve its goal (expressed in a formula of the language of the 
planning domain), the agent generates a plan using a classical planner (e.g., we used Fast-Forward \cite{hoffmann2001ff}),
and  then it executes the plan. However, the symbolic actions of the plan need to be translated into sequences of \emph{operations} executable by the agent's actuators (e.g., rotate of $30^\circ$, grasp the object in position $x,y,z$, move forward of 30cm). Designing effective and robust methods
for producing this mapping is a research area which goes out of the scope of this paper, see for instance
\cite{eppe2019semantics}. 
In our experiments, we adopt state-of-the-art path planning algorithms (based on a map
learned online by the agent) and ad-hoc compilations of actions. However, it is worth noting that 
we do not assume the execution of the actions leads to the symbolic state 
predicted by the planning domain. 
For instance, the execution of the action \code{Go\_Close\_To}($c$) might
end up in a situation where the agent is not close enough to the object $c$ and the predicate \code{Close\_To}($c$)
is false, despite being a positive effect of the action \code{Go\_Close\_To}($c$). Moreover, the execution of a symbolic action can have effects that are not predicted by the action schema.
For instance some properties of an object might become true even if they are not in the positive effects of
the symbolic actions.
For these reasons, after action executions, the agent must check
if the plan is still valid, and if not, it should react to the unexpected situation, e.g., by replanning. 

\section{Problem Definition}
We place an agent in a random position of an unknown environment; 
we initialize it with the following components:
$(i)$
a set of sensors on the environment;  
$(ii)$
a trained object detector $\rho_o$;
$(iii)$
a planning domain $\D=(\P,\OP,\H)$; 
$(iv)$
a method for executing its ground actions;
$(v)$
\emph{an untrained neural network}
$\rho_{t,p}$ for predicting the property $p$ of the objects of type
$t$, for a subset of pairs $(t,p)$ of interest.


We focus on the online training of the $\rho_{t,p}$'s; 
our aim is to design a general method to autonomously 
generate symbolic plans 
for producing a training set $T_{t,p}$, for every pair $(t,p)$ of interest, and use $T_{t,p}$ to train the perception function $\rho_{t,p}$.  

$T_{t,p}$ contains pairs $(c,v)$, where $c$ is (the name of) an object of type $t$ with the associated anchor (e.g., the visual features of the object) and $v\in\{p,\neg p\}$ is the value of the property $p$.
Since $T_{t,p}$ is automatically created by acting in the environment, 
it may contain wrong labels. We evaluate the effectiveness of our method on the performance (precision and recall) of each $\rho_{t,p}$ against a ground truth data set collected independently by the agent. 


\section{Method}
\label{sec:method}


We explain the proposed method with a simple example. 
%
%
Suppose an agent aims to
learn to recognize the property \code{Is\_Turned\_On} for objects
of type \code{Tv}, it can proceed as follows: 
\emph{(i)} look for an object
(say \code{tv$_0$}) of type \code{Tv}; \emph{(ii)} turn \code{tv$_0$}
on to make sure that \code{Is\_Turned\_On(tv$_0$)} is true,
\emph{(iii)} take pictures of \code{tv$_0$} from several perspectives,
and label them as positive examples for \code{Is\_Turned\_On}.
To produce negative examples for the same property, the agent can
proceed in the same fashion, applying the action \code{Turn\_Off(tv$_0$)}.

The behaviour explained above should be automatically produced and
executed by the agent for every learnable pair $(t,p)$, where $t$ denotes an object type and $p$ a learnable property.
Therefore, in the following, we explain a procedure that 
extends automatically the planning domain of the agent 
to express the goal of learning $p$ for $t$,
and such that the procedure for collecting training data for $\rho_{t,p}$
is generated by a symbolic planner, and can be executed by the agent. 
This method requires that, for every learnable pair $(t,p)$, the planning domain 
contains
at least an operator applicable to object of type $t$ that makes $p$ true,
and one that makes $p$ false.

%

%

%

%
This means that we have to extend 
the planning domain
with
the capability of expressing facts about its properties and types, i.e., we have to
extend it with meta predicates and names for the elements of the planning domain $\D$. 

\begin{table}\centering
\fbox{\parbox{\textwidth}{
\scriptsize
  \vspace{-2mm}
  \begin{tabbing}
    \hspace{20pt}\=\hspace{20pt}\=\kill 
    \texttt{Observe}$(o,t,p)$: \\
    \> $\prec$: \> $\neg$\texttt{Viewed}$(o,t,p)$ \\
    \>       \> \texttt{Closed\_To}(o)  \\
    \>       \> \texttt{Known}$(o,t,p)$ \\
    \> $\eff^+$: \> \texttt{Sufficient\_Obs}($t,p$) \\ 
    \> \> \texttt{Viewed}$(o,t,p)$\\ 
      \texttt{Explore\_for}$(t,p)$\\
    \> $\prec$:\>$\forall x(\texttt{Known}(x,t,p)\rightarrow \texttt{Viewed}(x,t,p))$ \\
    \>       \> $\lnot$ \texttt{Sufficient\_Obs}($t,p$)  \\
    \> $\eff^+$:\> \texttt{Explored\_for}$(t)$\\ 
  \texttt{Train}$(t,p,q)$: \\    
    \> $\prec$ \> \texttt{Sufficient\_Obs}($t,p$) \\ 
    \> \> \texttt{Sufficient\_Obs}($t,q$)\\
    \> $\eff^+$:\> \texttt{Learned}($t,p,q$)   
  \end{tabbing}  \vspace{-2mm}}}
  \caption{\label{tab:extended}Schemas for \code{Observe}, \code{Explore\_for} and \code{Train}.}
\end{table}


\subsection{Extended Planning Domain for Learning}

Table~\ref{tab:extended} summarizes how we extend the planning domain for observing, exploring and learning.

\paragraph{Names for types and properties}
For each object type $t\in\P$ (e.g., \code{Box}), we add a new constant \name{$t$} (e.g., \name{\code{box}})%
\footnote{Quotes are used to indicate names for elements of $\P$.}.
For each object property $p\in\P$ (e.g., \code{Is\_Open}),
  we add two new constants, namely \name{$p$} and \name{\code{not}$\_p$} (e.g., \name{\code{is\_open}}
  and \name{\code{not\_is\_open}}).

\paragraph{Epistemic predicates}
We extend $\P$ with predicates for stating that an agent knows/believes that an object has a certain property in a given state. The binary predicate \code{Known}($o$,\name{$p$}) (resp.  \code{Known}$(o,\name{\code{not}\_p})$) indicates that the agent knows that the object $o$ has (resp. does not have) the property $p$.  The atom \code{Known}($x$,\name{$p$}) is automatically added to the positive (resp. negative) effects of all the actions that have $p(x)$ in their positive (resp. negative) effects;
similarly, the atom \code{Known}($x$,\name{\code{not\_}$p$}) is automatically added to the positive (resp. negative) effects of all the actions that have $p(x)$ in their negative (resp. positive) effects. 
For example, the atoms \code{Known}($x$, \name{\code{is\_turned\_on}})
and \code{Known}($x$, \name{\code{not\_is\_turned\_on}}) are added to
the positive and negative effects of \code{Turn\_On($x$)}, respectively.
Similarly, the atoms \code{Known}($x$, \name{\code{is\_turned\_on}})
and \code{Known}($x$, \name{\code{not\_is\_turned\_on}}) are respectively added to
the negative and positive effects of \code{Turn\_Off($x$)}.

\paragraph{Predicates and Operators for Observations}
%

We extend the planning domain with the operator
\code{Observe}$(o,t,p)$, which takes as input an object $o$, a type
$t$, and a property $p$. The low level execution of \code{Observe}$(o,t,p)$
consists in extending the training dataset $T_{t,p}$ with observations (i.e., images) of object $o$ taken
from different perspectives. 
The positive effects of
\code{Observe}$(o,t,p)$ contain the atom \code{Viewed}$(o,p)$,
and the preconditions of \code{Observe}($o$,$t$,$p$) contain the atom $\neg\code{Viewed}(o,p)$,
which prevents the agent from again observing $o$ for the property $p$ in the future. 

The atom \code{Sufficient\_Obs}$(t,p)$ is added to the positive effects of the action \code{Observe}$(o,t,p)$.
%
%
%
%
Whether the agent, after executing \code{Observe}$(o,t,p)$, has not collected enough observations of objects of type $t$ with property $p$, 
the atom \code{Sufficient\_Obs}$(t,p)$ is actually false, in contrast with what is
predicted by the planning domain, 
and the agent has to plan for observing other objects of type $t$. 

\paragraph{Predicates and Operators for Exploration}
The planning domain is extended with the binary operator \code{Explore\_for}$(t,p)$ that explores the environment
looking for new objects of type $t$. 
The precondition of \code{Explore\_for}$(t,p)$ is that all the known objects of type $t$
have been viewed for the property $p$, i.e., $\forall x(\code{Knows}(x,t,p)\rightarrow \code{Viewed}(x,t,p))$. 
Indeed, finding a new object creates a new object $o$ in the planning domain, and makes aware the agent that  properties of $o$ can be observed.
%
%
\code{Explored\_for}($t$) is a positive effect of
\code{Explore\_for}$(t,p)$ in the planning domain. Such an effect indicates that the environment has been (even partially) explored for finding new objects of type $t$.
However, the actual execution of \code{Explore\_for}$(t,p)$ 
will not make it true until
 the environment has been completely explored, or a maximum number of iterations has been reached.
%


\paragraph{Predicates and Operators for Learning}
%
We extend the planning domain with the predicate \code{Learned}$(t,p,\code{not\_}p)$, indicating if the agent has collected enough observations, and trained $\rho_{t,p}$.
We add to the planning domain the operator \code{Train}$(t,p,q)$. 
When the agent executes the action
\code{Train}$(t,p,q)$, the network $\rho_{t,p}$ is trained using
$T_{t,p}$ as positive examples and $T_{t,q}$ as negative examples.
The preconditions
of this action include \code{Sufficient\_Obs}$(t,p)$ and
\code{Sufficient\_Obs}$(t,q)$ that guarantee to have
sufficient positive and negative examples for training $\rho_{t,p}$.  This action has only one positive effect,  which is 
\code{Learned}$(t,p,q)$.

\paragraph{Specifying the goal formula}
In the extended planning domain, the goal formula $g$ for learning a property $p$ for an object type $t$ is defined as:
\begin{align}
  \label{eq:goal-tp}
g &  = \code{Learned}(t, p, \code{not\_}p) \vee \code{Explored\_for}(t).
\end{align}
%
%
%
For example, suppose that an agent aims to learn the property \code{Turned\_On}
for objects of type \code{Tv}, then $g=\code{Learned(\name{tv},
  \name{turned\_on}, \name{not\_turned\_on})}\vee\code{Explored\_for{(\name{tv})}}$.
If the current set of constants contains an object, say \code{tv$_0$},
of type \code{Tv} such that 
\code{Viewed(tv$_0$,\name{tv},\name{is\_turned\_on})}
and \code{Viewed(tv$_0$,\name{tv},\name{not\_is\_turned\_on})} are both false,
then the goal is reachable by the plan:
\begin{center}
  \scriptsize
\begin{tabular}{l}
  \code{Go\_Close\_To(tv$_0$)}\\
  \code{Turn\_On(tv$_0$)}\\
  \code{Observe(tv$_0$, \name{tv},\name{turned\_on})} \\
  \code{Turn\_Off(tv$_0$)} \\
  \code{Observe(tv$_0$,\name{tv}, \name{not\_turned\_on})} \\
  \code{Train(\name{tv}, \name{turned\_on},\name{not\_turned\_on})}.
\end{tabular}
\end{center}
After the execution of all the actions but the last one of the above plan, if the agent has not collected enough training data 
for
\code{\name{turned\_on}} and \code{\name{not\_turned\_on}}, the atoms
\code{Sufficient\_Obs(\name{tv},\name{turned\_on})} and
\code{Sufficient\_Obs(\name{tv},\name{not\_turned\_on})} will be false,
and the last action of the plan cannot be executed. In such a case, the
agent has to replan in order to find another tv which has not been
observed yet.

Finally, notice that whether all the TVs known by the agent have been observed for the property \code{Turned\_On}, then the formula
$\forall x (\code{Known($x$,\name{tv},\name{turned\_on})}\rightarrow\code{Viewed}(x,\code{\name{tv}},\code{\name{turned\_on}})$
is true, and the 
goal can be achieved by generating a plan that satisfies \code{Explored\_for{(\name{tv})}},
i.e., by executing the action 
\code{Explore\_For(\name{tv},\name{turned\_on})}, which explores the environment for
new TVs. 

\subsection{Main control cycle}
%
  \begin{algorithm}[t]
  \caption{
\textsc{Plan and Act to Learn Object Props}\label{alg:palop}}
\begin{algorithmic}[1] 
\REQUIRE $\D=(\P,\O,\H)$ a planning domain
\REQUIRE $g = \bigwedge_{(t,p)\in TP}(\code{Learned}(t,p)\vee\code{Found\_New}(t))$
\STATE extend $\D$ with actions and predicates for learning
\STATE $\C\gets $ names for types and properties in $\P$\label{algline:init-c}
\STATE $s\gets \emptyset$\label{algline:init-s}
\STATE $T_{TP}\gets\{T_{t,p}=\emptyset\mid (t,p)\in TP\}$ \label{algline:initialize-T_TP}
\STATE $\rho_{TP}\gets\{\rho_{t,p}=\mbox{random init.}\mid (t,p)\in TP\}$ \label{algline:initialize-rho_TP}
\STATE $\pi\gets\textsc{Plan}(\D(\C),s,g)$ \label{algline:init_plan}
\WHILE{$\pi \neq \langle\rangle$} \label{algline:start-while}
    \STATE $op\gets\textsc{Pop}(\pi)$ \label{algline:get-first-action}
    \STATE $s \gets s \cup \eff^+(op) \setminus \eff^-(op)$ \label{algline:update-state-effects}      
    \STATE $\C, T_{TP},\rho_{TP}\gets \textsc{Execute}(op)$ \label{algline:exec}
    \STATE $s\gets\textsc{Observe}()$ \label{algline:update-state}
    \STATE $\pi\gets\textsc{Plan}(\D(\C),s,g)$ \label{algline:plan}
\ENDWHILE \label{algline:end-while}
\end{algorithmic}
\end{algorithm}

The main control cycle of the agent is described in Algorithm~
\ref{alg:palop}, which takes as input a planning domain $\D$ and the goal $g$ for learning a set $TP$ of type-property pairs.  
At the beginning, the set of constants $\C$ contains only the names for
types and properties, and the state $s$ is empty (lines \ref{algline:init-c}--\ref{algline:init-s}). For every pair $(t,p)\in TP$, the algorithm initializes
the training set $T_{t,p}$ to the empty set,
and the neural networks $\rho_{t,p}$
(lines \ref{algline:initialize-T_TP}--\ref{algline:initialize-rho_TP}).
Then, a plan $\pi$ is generated (line \ref{algline:init_plan}).
In the while loop (lines \ref{algline:start-while}--\ref{algline:end-while}), the state $s$ is updated according to
the action schema (line \ref{algline:update-state-effects}).
Next, the first action of the plan is executed and 
the set of known constants $\C$, the datasets $T_{t,p}$, and the neural
networks $\rho_{t,p}$ are updated (line \ref{algline:exec}).
Notice that, since the perceived effects of actions might not be consistent with those contained in the action schema, 
a sensing using the not
trainable perception functions is necessary, and the state is updated
accordingly (line \ref{algline:update-state}). Moreover, since $\pi$ might be no more valid in
the updated state, a new plan must be generated (line \ref{algline:plan}). 
The algorithm terminates if either the whole environment has been
explored or a maximum number of iterations has been reached, since, in such cases, the atom \code{Explored\_for}$(t)$ is set to true,
and plan $\pi$ for $g$ is empty.





\section{Experimental evaluation}
\label{sec:experiments}


%
We evaluate our approach on the task of collecting a dataset and training a set of neural networks to predict the
four properties \code{Is\_Open}, \code{Dirty}, \code{Toggled}, and \code{Filled} on 32 object types,
resulting in 38 pairs $(t,p)$, since not all properties are applicable to all object types. 

\paragraph{Simulated environment}
%
We experiment our approach in the ITHOR \cite{ai2thor}
photo-realistic simulator of four types of indoor environments: kitchens, living-rooms,
bedrooms, and bathrooms.
ITHOR simulates a robotic agent that navigates the environment and interacts with
the objects by changing their properties (e.g., opening a box, turning on a tv). 
The agent has two sensors: a position sensor and an on-board RGB-D camera.
For our experiment we split the 120 different environments, provided by ITHOR,
into 80 for training, 20 for validation, and 20 for testing. Testing environments are evenly distributed among the 4 room types.
\paragraph{Object detector} For the object detector $\rho_{o}$, we used the YoloV5 model \cite{yolov5_repo}, which takes as input an RGB image and returns the object types and bounding boxes detected in the input image.
For training $\rho_o$, we have generated the training (and validation) sets by randomly navigating in the training (and validation) environments, and using the ground truth object types and bounding boxes provided by ITHOR. The training and validation sets contain 115 object types and are composed by 259859 and 56190 examples, respectively.  
For validating the object detector, we performed 300 runs (with 10 epochs for each run) of the genetic algorithm proposed in \cite{yolov5_repo}.

\paragraph{Property predictors}
For the perception functions $\rho_{t,p}$ predicting properties, we adopted a ResNet-18 model \cite{he2016deep} with an additional fully connected linear layer, which takes as input the RGB image of the object and returns the probability of $p$ being true for the object. We consider that the input object has the property $p$ if the probability is higher than a given threshold (set to 0.5 in our experiments).

\paragraph{Evaluation metrics and ground truth}
%
We evaluate each trained $\rho_{t,p}$ using precision and recall
against a test set $G_{t,p}$, obtained by randomly navigating the 20 testing environments and using the ground truth information provided by ITHOR.  
%
%
In particular, for the \code{Is\_Open} property we generated a test set with 8751 examples, 2512 for the \code{Toggled} property, 1310 for the \code{Filled} property, and 3304 for the \code{Dirty} one. It is worth noting that the size of the test set for the \code{Is\_Open} property is higher than other ones since the number of object types that can be open is higher than the ones with other properties. 

\subsection{Experiments with the simulator}

We run our approach in each testing environment for training the neural network model associated to $\rho_{t,p}$ with the  training set $T_{t,p}$ collected online. 
At each run, the agent starts in a random position of the environment and executes 2000 iterations, where at each iteration a low-level operation (e.g., move forward of $30cm$) is executed.
%
%
%




To understand how the errors of the objects detector affect the performance, we propose two variants of our approach, namely \textsf{ND} (Noisy Detections) and \textsf{GTD} (Ground Truth Detections). In both variants, the agent trains $\rho_{t,p}$ on the training set $T_{t,p}$ collected in a single environment, and is evaluated on the test set $G_{t,p}$ previously generated in the same environment. In the \textsf{ND} variant, the agent is provided with a pre-trained object detector $\rho_{o}$; while in the \textsf{GTD} variant the agent is provided with a perfect $\rho_{o}$, i.e., the ground truth object detections provided by ITHOR.
In both variants, the neural networks $\rho_{t,p}$'s are trained for $10$ epochs with $1e^{-4}$ learning rate; the other hyperparameters are set to the default values provided by PyTorch1.9 \cite{paszke2019pytorch}. 

\begin{table}[t]\resizebox{0.45\textwidth}{!}{
    \begin{tabular}{lll|ll|ll|ll}  \toprule
& \multicolumn{2}{c|}{size of $G_{t,p}$}  &  \multicolumn{2}{c|}{size of $T_{t,p}$} &  \multicolumn{2}{c|}{Precision}&  \multicolumn{2}{c}{Recall}\\ \midrule
 Object type &    \textsf{ND} &    \textsf{GTD} &  \textsf{ND} &    \textsf{GTD} &  \textsf{ND} &    \textsf{GTD} &   \textsf{ND} &   \textsf{GTD}\\ \hline
  \multicolumn{9}{c}{\code{Dirty}} \\ \midrule 
         bed &      564 &         564 &     1502 &         671 &          0.95 &        0.57 &       0.43 &        0.61 \\
        bowl &      280 &         280 &      383 &        1027 &          0.67 &        0.98 &       0.81 &        0.73 \\
       cloth &       96 &         210 &       61 &         503 &          0.93 &        0.95 &       0.78 &         0.7 \\
         cup &       96 &         262 &      146 &         986 &          0.63 &        0.99 &       0.95 &        0.54 \\
      mirror &      654 &         678 &     2490 &        3100 &          0.91 &         0.9 &       0.68 &         0.8 \\
         mug &      230 &         432 &      225 &        1367 &          0.88 &        0.94 &       0.42 &        0.74 \\
         pan &      140 &         200 &       20 &         476 &          0.76 &        0.99 &       0.87 &        0.79 \\
       plate &      166 &         406 &       47 &        1304 &          0.61 &        0.97 &       0.97 &        0.77 \\
         pot &      210 &         272 &       51 &         929 &          0.76 &        0.99 &       0.91 &        0.98 \\
         \hline 
Weighted avg &        - &           - &        - &           - &          0.84 &        0.89 &       0.68 &        0.74 \\ \hline \hline
\multicolumn {9}{c}{\code{Filled}} \\ \hline
      bottle &       22 &          22 &       78 &         150 &          0.65 &           0 &          1 &           0 \\
        bowl &      328 &         256 &      390 &        1091 &          0.64 &           1 &       0.73 &        0.77 \\
         cup &      116 &         286 &      200 &        1028 &          0.92 &         0.9 &       0.56 &        0.68 \\
  houseplant &       34 &          34 &       18 &          72 &           0.5 &         0.5 &       0.65 &        0.82 \\
      kettle &      - &          84 &      - &         337 &           - &        0.25 &        - &         0.4 \\
         mug &      126 &         354 &      250 &        1136 &           0.8 &        0.86 &       0.51 &        0.56 \\
         pot &      226 &         274 &       93 &         809 &          0.67 &           1 &       0.89 &        0.79 \\
      \hline
Weighted avg &        - &           - &        - &           - &           0.7 &        0.86 &       0.72 &        0.66 \\ \hline \hline
  \multicolumn {9}{c}{\code{Is\_Open}} \\ \hline
         book &      148 &         268 &      367 &        1471 &             1 &        0.94 &       0.76 &        0.81 \\
          box &      204 &         204 &      959 &        1044 &          0.92 &        0.88 &       0.37 &        0.54 \\
      cabinet &     2892 &        2892 &     1545 &        1669 &          0.81 &         0.8 &       0.74 &        0.79 \\
       drawer &     3343 &        3747 &     1237 &        2624 &          0.79 &        0.75 &       0.77 &        0.71 \\
       fridge &      400 &         400 &      803 &        1109 &          0.78 &        0.81 &       0.72 &        0.75 \\
       laptop &      360 &         360 &     1124 &        1531 &          0.93 &        0.97 &       0.85 &        0.82 \\
    microwave &      250 &         250 &      742 &         843 &          0.68 &        0.82 &        0.5 &        0.68 \\
showercurtain &      144 &         134 &      271 &         567 &          0.47 &        0.96 &       0.41 &        0.76 \\
   showerdoor &       74 &         140 &       56 &         346 &          0.88 &        0.71 &       0.19 &        0.98 \\
       toilet &      356 &         356 &     1024 &        1148 &          0.89 &         0.9 &       0.63 &        0.74 \\
      \hline
 Weighted avg &        - &           - &        - &           - &          0.81 &         0.8 &       0.72 &        0.75 \\ \hline \hline
  \multicolumn {9}{c}{\code{Toggled}} \\ \hline
       candle &       54 &         124 &        3 &         118 &          0.59 &        0.33 &       0.63 &         0.6 \\
    cellphone &      - &         216 &      - &         682 &           - &        0.84 &        - &        0.94 \\
coffeemachine &      320 &         320 &      999 &         996 &          0.95 &        0.97 &       0.72 &        0.61 \\
     desklamp &       12 &          56 &      254 &         255 &             1 &        0.91 &          1 &        0.97 \\
      desktop &      - &          56 &      - &         184 &           - &           1 &        - &        0.93 \\
       faucet &      602 &         480 &      921 &        1663 &          0.84 &        0.85 &       0.89 &        0.92 \\
    floorlamp &       44 &          12 &       88 &          68 &          0.83 &        0.75 &        0.5 &           1 \\
       laptop &      432 &         432 &     1545 &        1777 &          0.91 &        0.83 &       0.61 &        0.74 \\
    microwave &      252 &         252 &     1131 &        1124 &             1 &           1 &       0.76 &        0.72 \\
   showerhead &      - &          46 &      - &          12 &           - &           1 &        - &           1 \\
   television &      222 &         238 &      269 &         510 &          0.99 &        0.94 &       0.85 &        0.95 \\
      toaster &      280 &         280 &      713 &        1072 &          0.86 &        0.98 &       0.59 &         0.7 \\
      \hline\hline
 Weighted avg &        - &           - &       - &           - &           0.9 &        0.88 &       0.74 &         0.8 \\
\bottomrule
\end{tabular}}
\caption{Size of the ground truth test set $G_{t,p}$, the generated training set $T_{t,p}$, and
  performance in terms of precision and recall on the 38 type-property pairs. 
\label{tab:metrics}}
\end{table}

\paragraph{Experimental results}
We compare the versions \textsf{ND} and \textsf{GTD} for each learned property; the results are shown in Table \ref{tab:metrics}.
In particular, the columns of Table \ref{tab:metrics} contain the object type, the number of examples collected in the training and test sets, respectively $G_{t,p}$ and $T_{t,p}$, the metrics precision and recall averaged over all 20 environments.
It is worth noting that the size of the test set can vary among \textsf{ND} and \textsf{GTD}, since we remove from the test set the object types that are missing in the training set, i.e., the object types that have not been observed by the agent. This is because we are interested in evaluating the learning performance on the object types that the agent actually manipulates and observes. 
Moreover, there are particular object types (e.g., desktop and showerhead in Table \ref{tab:metrics}) that are never recognized by the object detector, hence they are missing in the training set, and they are assigned the `-' value in Table \ref{tab:metrics}.   
%
%

Table \ref{tab:metrics} shows the results obtained for learning properties \code{Dirty}, \code{Filled}, \code{Is\_Open}, and \code{Toggled}. 
Not surprisingly, both the weighted average precision and recall of the \textsf{GTD} version are almost always higher than the \textsf{ND} ones, i.e., the overall learning performance are better when the agent is provided with ground truth object detections. 
%
The recall is generally lower than the precision, this is because for almost all object types, the number of negative examples is higher than the positive one, i.e., the training datasets are not balanced. Therefore, the agent is more likely to predict that a property is false, which causes more false negatives and a decrease of the recall. 
%
%
In our experiments, we tried to balance the observations in the collected training sets by randomly removing positive or negative examples, but we obtained worse performance. More sophisticated strategies might measure the information of each observation and remove the less informative ones; however, this problem is out of the scope of this paper.

\begin{figure}[t]
  \centering
  \includegraphics[width=0.37\textwidth]{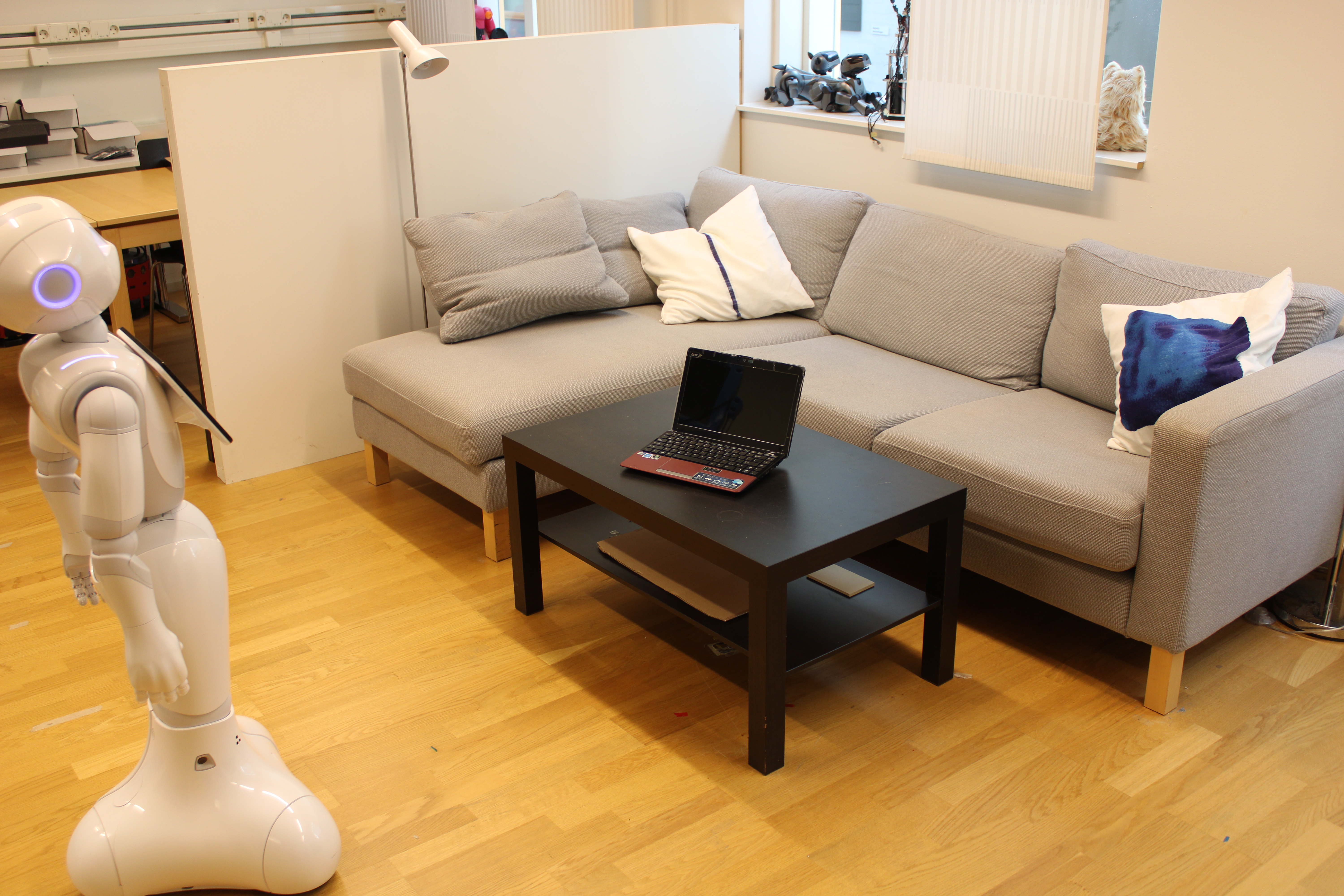}
  \caption{Pepper taking images of a laptop and asking a human to manipulate it for learning the property \code{Folded}.}
  \label{fig:peishome}
\end{figure}

%
In the \code{Dirty} property results, for all object types but bed, the number of examples in the training set is higher for \textsf{GTD}, as expected. The examples of beds in \textsf{GTD} are lower because in all bedrooms the agent focused on observing other object types. Indeed, for all other object types contained in bedrooms (i.e., cloth, mug and mirror), the examples collected by \textsf{GTD} are more than the \textsf{ND} ones. 
%

Moreover, for the \code{Dirty} property, the precision obtained by \textsf{GTD} is significantly higher than the \textsf{ND} one, for almost all object types (i.e., 7 out of 9).
For the mirror object type, the precision achieved by both \textsf{ND} and \textsf{GTD} is almost equal.
Remarkably, for the bed object type, the precision of the \textsf{GTD} version is much lower than the \textsf{ND} one. This is because, for large objects such as beds, the \textsf{GTD} version is more likely to collect examples not representative for the properties to be learned. For instance, the agent provided with ground truth object detections recognizes the bed even when it sees just a corner of the bed, whose image is not significant for predicting whether the bed is dirty or not.
Moreover, the examples of objects of type bed in the training set collected by \textsf{ND} is much higher than the \textsf{GTD} one.

The recall of the \textsf{GTD} version is not always higher than the \textsf{ND} one. In our experiments, we noticed that, for both \textsf{ND} and \textsf{GTD}, an high precision typically entails a low recall, and viceversa. This is because typically the agent collects more positive or negative examples of a single object type. For instance, the precision achieved by \textsf{ND} on object types bed, cloth, mirror and mug is high and the recall is low. Similarly, the recall achieved by \textsf{ND} on object types bowl, cup, pan, plate and pot is high and the precision is low. 
The recall obtained by \textsf{GTD} is lower than the precision for all object types but bed, where there is no significant difference.
Overall, the weighted average metric values show good performance, i.e., our approach is effective for learning to recognize properties without any dataset given a priori as input.
Similar considerations given for the \code{Dirty} property apply to results obtained for properties \code{Is\_Open}, \code{Toggled} and \code{Filled}, reported in Table \ref{tab:metrics}. 
However, it is worth noting that for the \code{Filled} property, the metric values obtained by both \textsf{ND} and \textsf{GTD} versions are particularly low for the object types houseplant and kettle. This is because, for the mentioned object types, the \code{Filled} property is hard to recognize from the object images. For instance, the fact that an object of type kettle is filled with water cannot be recognized from its image, since the water in the kettle is not visible from an external view such as the agent one.
Furthermore, \textsf{GTD} with the object type bottle achieves $0$ value of both precision and recall, this is a particular situation where the neural network associated to the \code{Filled} property never predicts false positives when evaluated on examples of objects of type bottle, hence precision and recall equals $0$.  

%
Finally, we compared the performance of the property predictors learned by \textsf{ND} (Table \ref{tab:metrics}) with a baseline where property predictors are trained on data manually collected as $G_{t,p}$. The baseline achieves an overall precision and recall of $0.81$ and $0.84$, respectively; while our approach achieves an overall precision and recall of $0.81$ and $0.71$, respectively. These results show that the online learning can achieve a precision comparable to the offline setting where data are manually connected, while its recall gets worse as the prediction gives false negative results more often.

\subsection{Real world demonstrator}



\begin{table}[t]
\centering
\resizebox{0.285\textwidth}{!}{
\begin{tabular}{ll|c|c}
\toprule

 Object type &  Property &  Precision &    Recall  \\
\midrule
        bowl & \code{Empty} &      0.63 &         0.98     \\
        laptop & \code{Folded} &       0.97 &         1.00     \\
        book & \code{Is\_Open} &      1.00 &         0.99     \\
        cup & \code{Filled} &      0.93 &         0.83     \\
      \hline\hline
 Weighted avg & - &  0.88 &           0.95              \\
\bottomrule
\end{tabular}
}
\caption{Precision and recall obtained by the neural networks predicting object properties in a real environment.}
\label{tab:PepperMetrics}
\end{table}

To test our method in a real-world setting, we used a Softbank's Pepper humanoid robot in PEIS home ecology \cite{saffiotti2005peis}, shown in Figure \ref{fig:peishome}. 
%
%
As an object detector, we adopted a publicly available model of YoloV5 pre-trained on the MS-COCO dataset \cite{lin2014microsoft}. 
For manipulation actions, Pepper asks a human to do the manipulations, due to its limited capabilities of manipulating objects. We used Pepper's speech-to-text engine for simple verbal interaction with the human. 
%
Given an object type and a property, Pepper first looks for the object and then asks the human about the property's state. Next, it collects samples and asks the human to change the state of the property, and after human confirmation, it further collects samples.

We run experiments for learning pairs (type, property) reported in Table \ref{tab:PepperMetrics}. 
%
%
For each pair, we run the experiment 7 times with different objects of the same type. 
%
%
At each run, Pepper collects 100 examples of the observed property, grouped into 50 positive and 50 negative examples. 
For each property, we took 4 runs for training (i.e., 400 examples), and 3 runs for testing (i.e., 300 examples).
Table \ref{tab:PepperMetrics} shows the precision and recall obtained on the test sets. Both the average precision and recall are high. For the simpler properties (i.e., \code{Is\_Open} and \code{Filled}), Pepper almost perfectly learned to recognize them. These results demonstrate that our approach can be effective also in real world environments.

%



\section{Conclusions and Future Work}

%
We address the challenge of using symbolic planning to 
automate the learning of perception capabilities. We focus on 
learning object properties, assuming that an object detector is pretrained. 
We experimentally show that our approach is feasible and effective. 
Still a lot of work must be done to address the general problem of planning and acting to learn in a physical environment. 
%
%
For example, planning for online training the object detector or learning relations among different objects.
%
We assume that actions that change an object property can be executed without being able to fully recognize the property itself. 
This is feasible in a simulated environment before deploying a robot in the real world. 
In a real world environment, where this assumption is more critical, we can use our method to improve agents' perception capabilities rather than learning them from scratch. 
Moreover, some actions can be executed without knowing the object properties, e.g., pushing a button to turn on a TV.

\section{Acknowledgements}

We thank the anonymous reviewers for their insightful comments. This work has been partially supported by AIPlan4EU and TAILOR, two projects funded by EU Horizon 2020 research and innovation program under GA n.\ 101016442 and n.\ 952215, respectively, and by MUR PRIN-2020 project RIPER (n.\ 20203FFYLK). 
This work has also been partially supported by the Wallenberg AI, Autonomous Systems and Software Program (WASP) funded by the Knut and Alice Wallenberg Foundation.

\bibliography{aaai23}

\end{document}